\documentclass[11pt]{article}
\usepackage{fullpage}

\usepackage[english]{babel}
\usepackage{xspace}
\usepackage{tcolorbox}
\usepackage{tikz}

\usepackage[colorlinks=true,
linkcolor=blue,
urlcolor=blue,
citecolor=blue]{hyperref}

\usepackage[utf8]{inputenc} 
\usepackage[T1]{fontenc}    
\usepackage{hyperref}       
\usepackage{url}            
\usepackage{booktabs}       
\usepackage{amsfonts}  
\usepackage{amsmath}
\usepackage{nicefrac}       
\usepackage{microtype}      
\usepackage{xcolor, colortbl}        
\usepackage{wrapfig}
\usepackage{caption}
\usepackage[capitalize,noabbrev]{cleveref}
\usepackage{tabularx}
\usepackage{subcaption}
\usepackage{footnote}
\makesavenoteenv{tabular}
\makesavenoteenv{table}
\usepackage{multirow}
\usepackage{bbm}
\usepackage{amsthm}
\usepackage[makeroom]{cancel}
\usepackage{stmaryrd}
\usepackage{amsmath,bm}
\usepackage{amssymb}
\usepackage{mathtools, cuted}
\usepackage{kantlipsum,setspace}
\usepackage{enumitem}
\usepackage{algorithm}
\usepackage{algorithmic}

\definecolor{Gray}{gray}{0.85}

\newtheorem{lemma}{Lemma}

\newtheorem{assumption}{Assumption}

\newcommand{\bth}{\boldsymbol{\theta}}

\title{From Demonstrations to Rewards: \\Alignment Without Explicit Human Preferences}

\author{\large Siliang Zeng$^{1}\footnote{\texttt{Email: zeng0176@umn.edu}}$, Yao Liu$^2$, Huzefa Rangwala$^2$, George Karypis$^2$, Mingyi Hong$^1$, Rasool Fakoor$^2$\\[.5cm]
	\small $^{1}$University of Minnesota, $^{2}$Amazon \\
    } 

\begin{document}

\maketitle

\begin{abstract}
One of the challenges of aligning large models with human preferences lies in both the data requirements and the technical complexities of current approaches. Predominant methods, such as RLHF, involve multiple steps, each demanding distinct types of data, including demonstration data and preference data. In RLHF, human preferences are typically modeled through a reward model, which serves as a proxy to guide policy learning during the reinforcement learning stage, ultimately producing a policy aligned with human preferences. However, in this paper, we propose a fresh perspective on learning alignment based on inverse reinforcement learning principles, where the optimal policy is still derived from reward maximization. However, instead of relying on preference data, we directly learn the reward model from demonstration data. This new formulation offers the flexibility to be applied even when \emph{only} demonstration data is available, a capability that current RLHF methods lack, and it also shows that demonstration data offers more utility than what conventional wisdom suggests. Our extensive evaluation, based on public reward benchmark, HuggingFace Open LLM Leaderboard and MT-Bench, demonstrates that our approach compares favorably to state-of-the-art methods that rely solely on demonstration data. Our code is available at \url{https://github.com/Hong-Lab-UMN-ECE/IRLAlignment}
\end{abstract}

\section{Introduction} 
\label{sec:intro}

Despite the success of aligning methods to human preferences~\cite{achiam2023gpt,team2023gemini,dubey2024llama,geminiteam2024gemini15unlockingmultimodal}, such as reinforcement learning from human feedback (RLHF) \cite{ziegler2019fine,ouyang2022training}, these approaches are quite complex and require various types of data. For instance, RLHF involves multiple stages: the first being supervised fine-tuning (SFT)~\cite{cen2025bridging}, which uses human demonstration data consisting of input prompts and their corresponding human response pairs. Next is reward modeling, which relies on preference data, where input prompts are paired with multiple responses that are ranked based on relative preference (e.g., the preferred vs. non-preferred response). The final step is policy learning through reinforcement learning (RL), where only input prompts are required, and generated responses are scored using the reward model learned in the previous step. These technical and data complexities make training such models challenging and costly, particularly due to the need for diverse data types like human demonstrations and preference annotations, which require extensive human input for accurate labeling and ranking. 

\begin{figure}[h]
\centering
\includegraphics[width=0.5\textwidth]{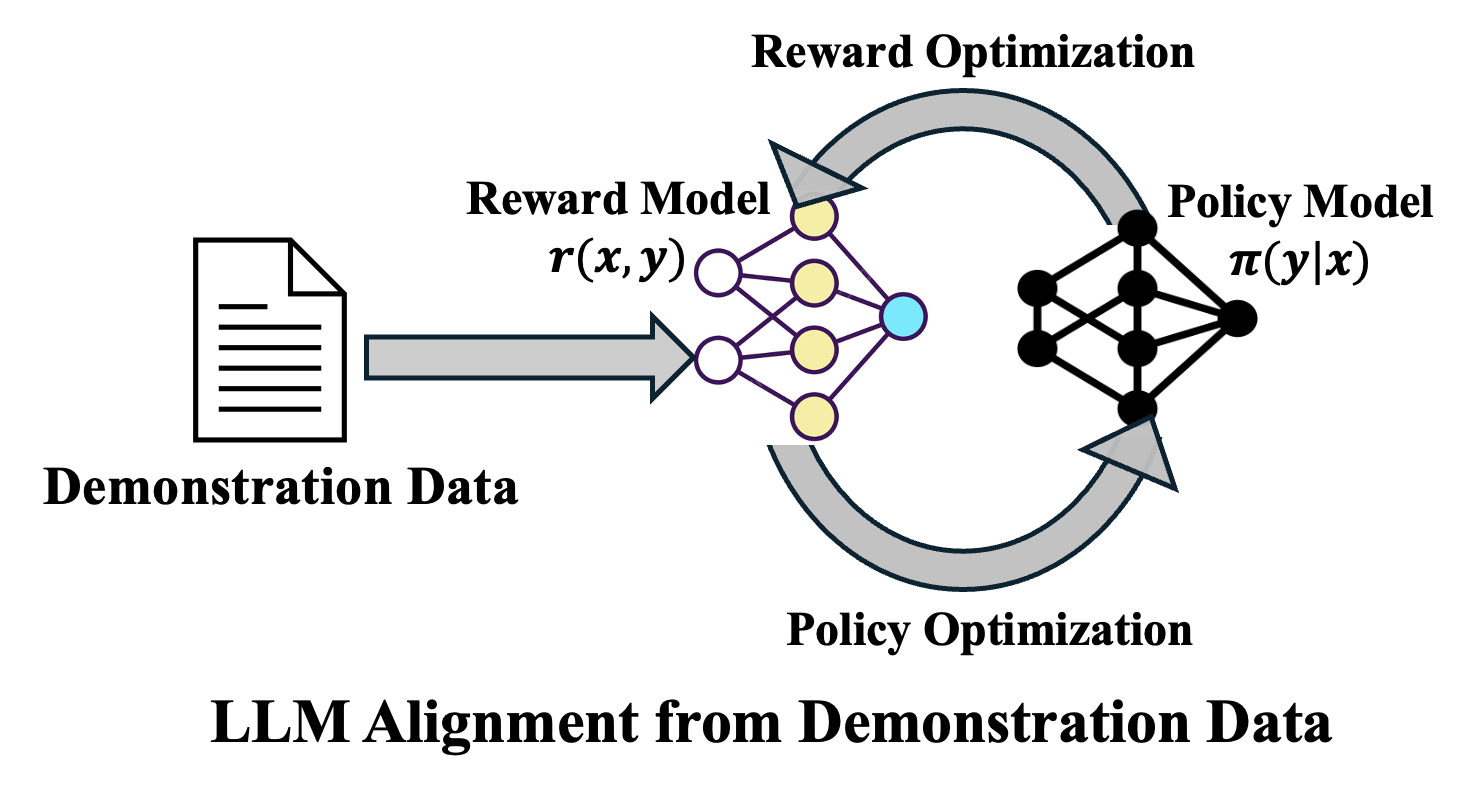}
\caption{Illustration of the iterative RLHF pipeline for LLM alignment from demonstrations through inverse reinforcement learning.}
\label{fig:memory}
\end{figure}

More specifically, one of the key challenges in RLHF is reward modeling, which acts as a proxy for human preferences. 
Existing approaches rely heavily on preference data to construct effective reward models demonstrating that a well-constructed reward function can boost model performance.~\cite{achiam2023gpt,team2023gemini,dubey2024llama}. Even direct preference alignment methods~\cite{zhao2023slichfsequencelikelihoodcalibration,an2023direct,rafailov2024direct,ethayarajh2024ktomodelalignmentprospect} which avoid explicit reward modeling, still necessitate preference data to align the model with human preferences. Considering the complexity of collecting high-quality preference data—and setting aside the modeling challenges—the key question is: Can we extract preferences from demonstration data alone, given that { such data is known to contain valuable human preference information}? 
One potential answer lies in the utilization of inverse reinforcement learning (IRL) to formulate this problem, which has the capability to learn both reward and policy simultaneously~\cite{ng2000algorithms,ziebart2008maximum,fu2017learning,zeng2022maximum}. This allows it to potentially learn preferences \emph{directly} from demonstration data, even when relying solely on that data. Motivated by this observation, we propose a bi-level formulation based on IRL for the alignment problem, where the reward model and policy are modeled separately but learned in an interleaved manner. The interleaving between policy learning and reward learning helps to model preferences solely from demonstration data, although those preferences are implicit in the data (see \cref{fig:memory}). As our results show, this formulation indeed helps to improve the performance of the final model, particularly when compared to SFT, which also relies solely on demonstration data.

{\bf Summary of contribution of this work:} 
\begin{itemize}[leftmargin=*]
\item We develop a new IRL-based method for alignment that relies solely on demonstration data, yet still improves the resulting model's performance. This method adopts a bi-level formulation, where the reward and policy are modeled separately. In this formulation, the policy is framed as the optimal solution to a KL-Divergence regularized policy optimization problem, constrained by a reward estimator. The reward model is optimized using the demonstration data to ensure that its corresponding optimal policy is the maximum likelihood estimator derived from the same dataset.

\item We also present a comprehensive comparison with a recently proposed method SPIN \cite{chen2024self}, which can demonstrate that our proposed method can outperform SPIN in both reward learning and policy learning through have separate parameterization for the reward model and policy model.

\item We demonstrate that the reward learned solely from demonstration data exhibits strong generative capabilities in assessing data quality. This is evidenced by evaluating the reward's accuracy on a hold-out preference dataset and a public reward benchmark ~\cite{lambert2024rewardbench}.

\item Empirically, we extensively evaluate our proposed method by fine-tuning the 1B Pythia model on the TL;DR dataset~\cite{stiennon2020learning,huang2024n+} and the 7B Mistral model on the UltraChat dataset~\cite{ding2023enhancing}. Our numerical results demonstrate that the proposed method compares favorably to both SFT and SPIN, as indicated by higher win rates in evaluations conducted by ChatGPT and improved performance on downstream tasks in the Open LLM Leaderboard~\cite{myrzakhan2024open} and MT-Bench~\cite{zheng2023judging}. The proposed method achieves the state-of-the-art performance compared with the demonstration-only alignment baselines.
 
\end{itemize}

\section{Related Work}
\paragraph{Imitation learning} Imitation learning assumes the availability of a demonstration dataset containing expert data and focuses on learning from these demonstrations to match the expert's policy. Behavior cloning is one classic imitation learning algorithm which directly fits the demonstration data through supervised learning \cite{pomerleau1988alvinn}. However, naively fitting sequential demonstrations can lead to distribution shift between demonstration trajectories and policy rollout. Moreover, since imitation learning where future transitions depend on
previous actions violates the common i.i.d. assumptions made in statistical learning, naively fitting demonstration trajectories can incur unfavorable regret bound which has quadratic dependence on the problem horizon \cite{ross2010efficient,ross2011reduction}. To address these challenges in imitation learning, it has been proposed to model the policy as the (optimal) solution to a MDP under a specific reward function \cite{osa2018algorithmic}. To learn a policy that effectively utilizes demonstration trajectories to match the expert policy, inverse reinforcement learning methods are proposed to search for one optimal reward estimator which can classify demonstration trajectories from other policy rollouts and guarantee that its corresponding optimal policy can imitate the observed expert behaviors in the demonstration dataset \cite{ziebart2008maximum,ho2016generative,fu2017learning,zeng2022maximum,garg2021iq}.

\paragraph{Imitation learning for language modeling}
The connection between imitation and generative language modeling can be traced back to some adversarial training methods for text generation~\cite{yu2017seqgan,wu2021textgail}, although some of them may not draw explicit connection with imitation learning. There are also works of applying inverse RL methods to a specific text generation tasks, such as table-to-text generation~\cite{ghosh2021helpful}, program generation~\cite{ghosh2021mapping}, and summarization~\cite{fu2022inverse}. To efficiently imitate the demonstration data, \cite{chen2024self} proposed an imitation learning algorithm, SPIN, which construct synthetic preference data through pairing demonstration and model-generated continuations and then leverage direct preference method \cite{rafailov2024direct} for policy fine-tuning. \cite{li2024getting} shows the connection between SPIN and IRL from a theoretical perspective and reveals that SPIN utilizes the parameterization technique developed in \cite{rafailov2024direct} to avoid explicit reward modelling and reward learning from computational simplicity. Although the parameterization technique introduced by \cite{rafailov2024direct} allows skipping the reward modeling subroutine and reduces the complexities of RLHF, it faces challenges due to the distribution shift between preference data and model outputs. This shift can lead to instability during the training process~\cite{xu2024dpo,ivison2024unpacking}. It is wroth noting that while \cite{li2024getting} discussed that the demonstration data can potentially benefit reward learning, it lacks practical algorithm for IRL methods which can iteratively enhance both reward model and policy by learning directly from demonstration data. More recently, ~\cite{cundysequencematch} and \cite{wulfmeier2024imitating} apply the imitation learning algorithm IQLearn for the post-training step of large language models, making their approaches closer to ours. In contrast, we adopt the framework of maximum likelihood inverse RL (ML-IRL) and derive a different objective, with results on broader LLM post-training benchmarks. Similar with us, \cite{sun2024supervised} draws a connection between inverse RL and the supervised fine-tuning problem of LLMs, but lacks both algorithmic implementation and experimental studies.

\section{Preliminaries}
We formulate the alignment problem of auto-regressive language models as a Markov decision process (MDP) following~\cite{ouyang2022training}. For a language model $\pi$, we denote its probability of generating a completion as $\pi(\bm{y} | \bm{x})$, where $\bm{x} = [x_1, x_2, \cdots, x_n] $ denotes the sequence of tokens in the input prompt and $\bm{y} = [y_1, y_2, \cdots, y_H]$ denotes the sequence of tokens in the model generated continuation. The language model generates each token auto-regressively in a sequential manner as
$
    \pi(\bm{y}|\bm{x})=\prod_{h=1}^{H}\pi(y_h|\bm{x},\bm{y}_{1:h-1}) 
$, where each step $h$ is viewed as a time-step in the MDP.


The current predominant method for aligning models with human preferences is RLHF~\cite{ouyang2022training}, which comprises mutiple stages, as outlined earlier. In the first stage, SFT, a high-quality human demonstration dataset \(\mathcal{D} = \{ (\bm{x}^i, \bm{y}^i) \}_{i=1}^N\) is used to fine-tune the pre-trained model using the following maximum likelihood objective:
\begin{equation}\label{eq:SFT}
\min _{\phi} \ \ell_{\mathrm{SFT}}(\phi):=-\mathbb{E}_{(\bm{x},\bm{y})\sim \mathcal{D}}\left[\log \pi\left( \bm{y} | \bm{x}; \phi \right)\right].
\end{equation}
In the reinforcement learning literature, this method is also known as behavior cloning \cite{osa2018algorithmic}. Notably, theoretical analyses of behavior cloning \cite{ross2010efficient,ross2011reduction} indicate that directly fitting sequential demonstration data in a MDP can result in unfavorable regret bounds, exhibiting quadratic dependence on the problem horizon. 
These insights suggest that SFT, as a form of behavior cloning, may not be the most effective approach for learning from demonstration in MDPs. To improve policy learning from sequential demonstrations, methods such as imitation learning and IRL \cite{ng2000algorithms,ho2016generative,osa2018algorithmic,zeng2022maximum} have been proposed, offering superior performance compared to naive behavior cloning methods.

In the second stage, there are two prominent classes of RLHF algorithms. One is explicitly building a parametric reward model and then fine-tuning the policy with online RL methods and the other is directly learning a policy from preference data. We refer to them as reward-based and reward-free methods in this paper. Reward-based RLHF approaches first train a reward model $r(\bm{x}, \bm{y}; \theta)$ by separating the score between preferred completion and non-preferred completion in a preference dataset $\mathcal{D}_{\mathcal{P}}:=\{(\bm{x},\bm{y}_w,\bm{y}_l)\}$ where $\bm{y}_w$ is {preferred} one over $\bm{y}_l$ according to the annotation from human annotator. (see e.g., \cite{christiano2017deep,stiennon2020learning, ouyang2022training}). More specifically, RLHF methods follow the Bradley-Terry model~\cite{bradley1952rank} which assumes that the distribution of preference label under one reward model $ r(\bm{x}, \bm{y})$ is represented as $\mathbb{P}\left(\bm{y}_w \succ \bm{y}_l \mid \bm{x} \right) =\sigma \big(r(\bm{x}, \bm{y}_{w}) - r(\bm{x}, \bm{y}_{l}) \big)$, where $\sigma(\cdot)$ is the sigmoid function. Therefore one can derive the reward learning objective from the maximum log-likelihood (MLE) on Bradley-Terry model:
\begin{align}\label{eq:BTL_loss}
    &\max_{\theta}\ \ell_{\rm RM}(\theta) \nonumber \\
    &:=\mathbb{E}_{(\bm{x}, \bm{y}_w, \bm{y}_l) \sim \mathcal{D}_{\mathcal{P}}} \Bigg[ \log \Big( \sigma \big( r(\bm{x}, \bm{y}_{w}; \theta) - r(\bm{x}, \bm{y}_{l}; \theta) \big) \Big) \Bigg]. 
\end{align}

After learning the reward model, various of online RL approaches can be used to fine-tune the policy from its own generation, for examples proximal policy optimization (PPO) \cite{schulman2017proximal}, variants of REINFORCE \cite{ahmadian2024back,li2023remax} and reward-ranked fine-tuning \cite{dong2023raft}. The most commonly used objective in this stage is the following KL-regularized reward maximization:
\begin{align}\label{eq:RLFT}
\max _{\pi} \ \ell_{\mathrm{RL}}(\pi)&:=\mathbb{E}_{\bm{x} \sim \mu, \bm{y} \sim \pi(\cdot|\bm{x})}\left[r(\bm{x}, \bm{y}; \theta)\right]  \nonumber \\
& \quad - \beta \mathbb{E}_{\bm{x} \sim \mu}[D_{\mathrm{KL}}(\pi\left(\cdot | \bm{x}\right)\|\pi_{\mathrm{ref}}\left(\cdot | \bm{x}\right))],
\end{align}
where $\pi_{\mathrm{ref}}$ is a fixed reference model (usually the SFT model) and $\mu(\cdot)$ denotes the prompt distribution over one prompt dataset. Here, in the policy optimization problem, the regularization of the KL divergence between the policy model $\pi$ and reference model $\pi_{\rm ref}$ ensures that the langugae model will not deviate from the reference model too much. The advantage of RLHF over SFT observed by \cite{stiennon2020learning,ouyang2022training,team2023gemini} comes from two aspects hypothetically: the generalization ability from reward model and learning on self-generated sequence. 

Reward-free RLHF approaches~\cite{rafailov2024direct,an2023direct} is an alternative to the classical reward-based RLHF by shortcutting the reward learning step, or implicitly learning it together with the policy learning. As an example, Direct Preference Optimization (DPO)~\cite{rafailov2024direct} propose to incorporate reward learning implicitly by utilizing the structure of the optimal solution of the RL problem in \cref{eq:RLFT}.
Based on that, DPO derives its objective as below:
\begin{equation}
\begin{aligned}
    \mathbb{E}_{(\bm{x}, \bm{y}_w, \bm{y}_l) \sim \mathcal{D}_{\mathcal{P}}} \Bigg[ \log \Big( \sigma \Big( &\beta\log\left( \frac{\pi(\bm{y}_w|\bm{x})}{\pi_{\text{ref}}(\bm{y}_w|\bm{x})} \right) - \beta\log\left( \frac{\pi(\bm{y}_l|\bm{x})}{\pi_{\text{ref}}(\bm{y}_l|\bm{x})} \right) \Big) \Big) \Bigg]. \label{eq:DPO_loss}
\end{aligned}
\end{equation}

\section{Problem Formulation} \label{sec:problem_formulation}

As we have mentioned in \cref{sec:intro} and also motivated by the theoretical understanding developed in \cite{ross2010efficient,ross2011reduction}, SFT or equivalently behavior cloning can incur unfavorable error bound which has quadratic dependence on the problem horizon when learning from demonstration data with sequential structure. To bridge the gap between the current SFT method and imitation learning methods in RL literature \cite{ng2000algorithms,ho2016generative,osa2018algorithmic,zeng2022maximum}, we consider a maximum likelihood formulation for IRL which this approach allows for the learning of a reward model and fine-tunes the SFT model using demonstration data.


\subsection{A Maximum Likelihood Formulation for Reward Learning and Policy Fine-tuning}

Given a demonstration dataset $\mathcal{D}$, the challenge lies in learning a reward model that aligns its corresponding policy with the demonstration data and effectively captures  the implicit human preferences contained within it. Unlike the standard RLHF, where the reward model is trained on a dataset of pairwise comparisons that explicitly represents human preferences, learning rewards from a demonstration dataset presents additional challenges due to the lack of explicit preferences in this data. Instead, this dataset contains only implicit preferences. 
Motivated by inverse reinforcement learning (IRL) based approaches \cite{ziebart2008maximum,ziebart2013principle,fu2017learning,zeng2022structural,zeng2022maximum}, we propose an IRL formulation grounded in maximum likelihood estimation to align the model with the demonstration dataset. 
Our maximum likelihood formulation aims to learn an ''optimal'' reward model such that its corresponding policy serves as the maximum likelihood estimator over the demonstration dataset. Here, we present our proposed formulation as follows:

\begin{subequations} \label{ML:formulation}
\begin{align}
	\max_{\theta} 	&~~~~L(\theta) := \mathbb{E}_{\bm{x} \sim \mu( \cdot), \bm{y} \sim \pi^{\rm E}(\cdot | \bm{x} ) } \big[\log \pi^*_{r_{\theta}}(\bm{y} | \bm{x} ) \big] \label{eq:ML} \\
	s.t &~~~~ 
\pi^*_{r_{\theta}} := \arg \max_{\pi} ~ \mathbb{E}_{\bm{x} \sim \mu( \cdot), \bm{y} \sim \pi(\cdot | \bm{x} )} \bigg[ r(\bm{x}, \bm{y}; \theta) - {D}_{\rm KL}\Big( \pi(\cdot | \bm{x}) \| \pi_{\text{ref}}(\cdot | \bm{x}) \Big) \bigg], 
\label{def:inner_problem}
\end{align}
\end{subequations}

where $\mu(\cdot)$ denotes the distribution of the prompt, $\pi$ denotes a policy model which generates continuations from given prompts, $ \pi^{\rm E} $ denotes the expert-level policy which can generates high-quality demonstration continuations and $\pi_{\rm ref}$ denotes one reference model which is usually chosen as the SFT model in LLM alignment. Moreover, $r(s,a; \theta)$ is the parameterized reward model and $ \pi^*_{r_{\theta}} $ denotes the optimal policy to a KL Divergence regularized policy optimization problem when the reward model is $r(\cdot, \cdot; \theta)$.


We now make some remarks about our maximum likelihood formulation of our alignment algorithm. First, the problem takes the form of a {\it bi-level} optimization problem, where the {\it upper-level} problem \cref{eq:ML} optimizes the reward parameter $\theta$, while the {\it lower-level} problem \cref{def:inner_problem} describes the corresponding policy $\pi^*_{r_{\theta}}$ as the solution to an KL Divergence regularized policy optimization problem \cite{stiennon2020learning,ouyang2022training}. As a remark, despite that our maximum likelihood formulation \cref{ML:formulation} establishes one framework to imitate the expert policy and estimate the reward model from demonstration dataset, it is impractical to continuously sample demonstration generations from the expert policy $\pi^{\rm E}$ in \cref{eq:ML} since the expert policy is unknown and only one observed demonstration dataset is available. To resolve this issue, we instead replace the expert policy by one fixed demonstration dataset which contains finite samples. 

Given a demonstration dataset $\mathcal{D} := \{ (\bm{x}, \bm{y}) \}$, we propose a surrogate objective $ \widehat{L}(\theta; \mathcal{D}) $ which approximates the maximum likelihood formulation \cref{ML:formulation} with finite demonstration data. Here, we consider the following surrogate problem:
{\small
    \begin{align}
	&\max_{\theta} 	~ \widehat{L}(\theta; \mathcal{D}) :=
 \mathbb{E}_{(\bm{x}, \bm{y}) \sim \mathcal{D}} \big[ r(\bm{x}, \bm{y}; \theta) + \log \pi_{\rm ref}(\bm{y}|\bm{x}) \big] - \mathbb{E}_{{\bm{x} \sim \mu( \cdot), \bm{y} \sim \pi^*_{r_{\theta}}(\cdot | \bm{x} )}}\bigg[ r(\bm{x}, \bm{y}; \theta) -  {D}_{\rm KL}\Big( \pi^*_{r_{\theta}}(\cdot | \bm{x}) \| \pi_{\text{ref}}(\cdot | \bm{x}) \Big) \bigg] \label{ML:estimation:surrogate} 
\end{align}
where the policy $ \pi^*_{r_{\theta}} $ denotes the optimal policy corresponding to the policy optimization problem defined in \cref{def:inner_problem} when the reward model is parameterized by the parameter $\theta$.

Based on the surrogate estimation problem \cref{ML:estimation:surrogate}, we show that the IRL for LLM alignment problem defined in \cref{ML:formulation} can be accurately approximated with a {\em finite} set of high-quality generations when the ``expert-level'' generative model (or data source) $\pi^{\rm E}$ is not known. In particular, 
below we show under a mild assumption about the boundedness of the reward score and the reference model, $\widehat{L}(\theta;\mathcal{D})$ can well-approximate $L(\theta)$ when the offline demonstration dataset includes sufficient number of high-quality generations.
\begin{assumption} \label{assumption:bound_reward}
For any reward parameter $\theta$, the following condition holds:
        \begin{align}
          0 \leq r(\bm{x},\bm{y};\theta) \leq C_r, \; C_p \leq \log \pi_{\rm ref}(\bm{y}|\bm{x}) < 0, \; \forall \bm{x},\bm{y}\label{ineq:bound_reward} 
        \end{align}
        where $C_r>0$ and $C_p < 0$ are fixed constants.
\end{assumption}


\begin{lemma} \label{lemma:objective_concentration}
    Suppose Assumption \ref{assumption:bound_reward} hold. Consider the likelihood function $L(\theta)$ in \cref{eq:ML} and its surrogate empirical version $\widehat{L}(\theta; \mathcal{D})$ defined in  \cref{ML:estimation:surrogate}. Then, with probability greater than $ 1 - \delta$, we have:
    \begin{align}
        |L(\theta) - \widehat{L}(\theta; \mathcal{D})| \leq (C_r - C_p) \sqrt{\frac{\ln(2 / \delta)}{2|\mathcal{D}|}}. \label{eq:approximation_error:concentration}
    \end{align}
\end{lemma}
The proof of Lemma \ref{lemma:objective_concentration} can be found in Appendix. 

\section{The Proposed Algorithm} 
We are now ready to design algorithms for the proposed maximum likelihood formulation \cref{ML:formulation} aimed at aligning large language models (LLMs) using demonstration data. To begin with, first note that the maximum likelihood formulation \ref{ML:formulation} and its surrogate estimation problem \cref{ML:estimation:surrogate} takes a hierarchical form, and it belongs to the class of problem named {\it bi-level} optimization. Generally speaking, bi-level problems are not easy to optimize since they have a nested structure for two optimization problems. In this section, we will propose one computationally tractable algorithm for both reward learning and policy fine-tuning to solve the LLM alignment problem \ref{ML:estimation:surrogate}.

Before presenting the details, we emphasize that throughout this section, we aim to explicitly identify {\it both} an optimal policy $\pi^*_{r_{\theta}}$ and a corresponding reward estimate $r(\cdot, \cdot;\theta)$ that align with the demonstration data. Specifically, the policy $\pi^*_{r_{\theta}}$ is considered an optimal solution with respect to the reward estimate $r(\cdot, \cdot;\theta)$, as defined by the policy optimization problem in equation \cref{def:inner_problem}. Given this optimal policy constraint relative to a specific reward estimate, we propose an algorithm to tackle this single-stage, bi-level problem. It is important to note that our approach is a departure from popular methods like DPO \cite{rafailov2024direct}, which directly optimize a fixed loss function (see \cref{eq:DPO_loss}) without explicitly modeling the reward. 

Our algorithm operates by alternating between two key steps: 1) {\bf Policy Alignment Step}, where we perform a policy improvement step towards solving \cref{def:inner_problem} under a fixed reward function $r(\cdot, \cdot;\theta)$, effectively aligning the policy with the current reward estimate, 2) {\bf Reward Alignment Step}, where we update the reward parameters $\theta$ using a stochastic gradient estimator to align the reward model with the demonstration dataset. This training loop allows the policy and reward model to be fine-tuned iteratively, potentially leading to better alignment with the demonstration data compared to standard SFT methods which treat the loss function \cref{eq:SFT} as fixed. 

In the following sections, we delve into each of these steps in greater details, providing theoretical insights and practical implementation considerations.



\textbf{Policy Alignment Step.} 
From our earlier discussion, we know that the optimal policy $\pi^*_{r_{\theta}}$ corresponds to the optimal solution to the policy optimization problem \cref{def:inner_problem} under a fixed reward model $r(\cdot, \cdot; \theta)$. To tackle such policy optimization problem, one can adopt the standard approaches such as the well-known proximal policy optimization (PPO) \cite{schulman2017proximal} algorithm to obtain an approximate optimal policy. As a remark, due to some practical difficulties for implementing PPO to fine-tune LLMs (like heavy memory cost and laborious hyper-parameter
tuning), it is possible to consider a simpler method than running PPO to obtain the optimal policy. Some of recently proposed RLHF methods like REINFORCE-type variants \cite{li2023remax,ahmadian2024back} and reward ranked fine-tuning \cite{dong2023raft} provides more computationally tractable alternatives to PPO for fine-tuning LLMs under one estimated reward model. It is important to note that, the point of the above discussion is that all of these different choices for policy optimization methods can be incorporated into our policy alignment step. 

From a theoretical perspective, based on \cite{cen2022fast,zeng2022maximum,ji2024self}, one can perform a ``soft policy iteration'' to obtain one updated policy estimator $\pi_{k+1}$ in each iteration $k$ as below:
\begin{align}
    \pi_{k+1}(\bm{y} | \bm{x}) \propto \pi_{\rm ref}(\bm{y} | \bm{x})\exp\big( r(\bm{x}, \bm{y};\theta) \big), \quad \forall s \in \mathcal{S}, a \in \mathcal{A}. \label{def:soft_policy_iteration}
\end{align}
To approximate the closed-form optimal policy in practice, one can utilize the popular policy optimization pipeline for fine-tuning LLMs under one reward model \cite{schulman2017proximal,ahmadian2024back}.
 

\begin{algorithm}[t] 
\caption{{\it Joint Reward Learning and Policy Fine-tuning from Demonstrations}} 
\begin{algorithmic}
\STATE {\bfseries Input:} A demonstration dataset $\mathcal{D}$, a reference model $\pi_{\rm ref}$ and $K$ the number of iterations.
\FOR{$k=0,\ldots, K-1$}
\STATE \textbf{{Policy Alignment:}} Run policy optimization subroutine (like PPO) to update the policy: $$\pi_{k+1}(\bm{y}|\bm{x})\propto \pi_{\rm ref}(\bm{y}|\bm{x}) \exp(r(\bm{x}, \bm{y};\theta_{k}))$$\\
\STATE \textbf{Reward Alignment:} Construct synthetic preference data and optimizing the following problem:
{\small
\begin{align}
    \theta_{k+1} := \arg \min_{\theta} -\mathbb{E}_{(\bm{x}, \bm{y}) \sim \mathcal{D}, \bm{y}^{\prime} \sim \pi_{k+1}(\cdot | \bm{x})} \Big[ & \log \Big( \sigma \big( r(\bm{x}, \bm{y}; \theta) - r(\bm{x}, \bm{y}^{\prime}; \theta) \big) \Big) \Big] \nonumber
\end{align}
}
\ENDFOR
\STATE \textbf{Output:} Estimated Reward Model $r(\cdot,\cdot;\theta_{K})$ and Policy Model $\pi_{K}(\cdot|\cdot)$
\end{algorithmic}
\label{alg:IRL_Alignment}
\end{algorithm}

{\bf Reward Optimization Step.}
We propose to use a stochastic gradient-type algorithm to optimize $\theta$. Towards this end, let us first 
 derive the exact gradient $\nabla \widehat{L}(\theta; \mathcal{D})$. See Appendix for detailed proof. 
\begin{lemma} \label{lemma:outer_gradient}
	The gradient of the finite-sample surrogate objective function $ \nabla \widehat{L}(\theta; \mathcal{D}) $ can be expressed as follows:
	\begin{align}
		\nabla \widehat{L}(\theta; \mathcal{D}) & = \mathbb{E}_{(\bm{x}, \bm{y})\sim \mathcal{D}}\big[ \nabla_{\theta} r(\bm{x}, \bm{y}; \theta) \big] -\mathbb{E}_{\bm{x} \sim \mu(\cdot), \bm{y}\sim \pi^*_{r_{\theta}}(\cdot | \bm{x})}\big[ \nabla_{\theta} r(\bm{x}, \bm{y}; \theta) \big]. \label{eq:likelihood_grad}
	\end{align}
\end{lemma}



To obtain stochastic estimators of the exact gradient $\nabla \widehat{L}(\theta; \mathcal{D})$, we take two approximation steps: 1) approximate the optimal policy  $\pi_{k+1}$ in \cref{def:soft_policy_iteration} through running a finite policy optimization steps in the RL subroutine since repeatedly estimating the optimal policy under each reward estimator can lead to computational burden; 2)  sample one batch of demonstration data from the demonstration dataset $\mathcal{D}$; 3) sample model-generated data from the current policy estimator. Theoretically, as long as the policy estimator achieves policy improvement in each IRL iterations, the training pipeline can be stable in such approximation and converge to the optimal solution \cite{zeng2022structural}.

Intuitively, following the reward gradient expression as shown in \cref{eq:likelihood_grad}, if the model-generated data from the current policy $\pi_{k+1}$ has not matched the demonstration dataset $\mathcal{D}$ yet, then the reward score should be improved by going towards the direction suggested by the demonstration data, while {\it going away} from those generated by the current policy. Similar to the BTL model, from the gradient expression \cref{eq:BTL_loss}, it is clear that the algorithm will find the reward update direction that increases the gap between the reward of the real samples (demonstrations) and the synthetic ones (model-generated continuations). Hence, as for each reward optimization step at iteration $k$, one can construct the following loss function to update the reward parameter $\theta$ through constructing synthetic preference data through pairing the demonstration data with the model generations:
{\small
\begin{align}\label{eq:IRL_RM_loss}
    & \min_{\theta}\ L_{\rm RM}(\theta; \mathcal{D}) :=-\mathbb{E}_{(\bm{x}, \bm{y}) \sim \mathcal{D}, \bm{y}^{\prime} \sim \pi_{k+1}(\cdot | \bm{x})} \Big[ \log \Big( \sigma \big( r(\bm{x}, \bm{y}; \theta) - r(\bm{x}, \bm{y}^{\prime}; \theta) \big) \Big) \Big]. \nonumber  
\end{align}
}

In summary, the proposed algorithm for solving \cref{ML:estimation:surrogate} is given in Alg. \ref{alg:IRL_Alignment}.


\section{Experimental Results}

In this section, we evaluate the effectiveness of our proposed method through comprehensive experiments on two distinct datasets: the TL;DR dataset \cite{stiennon2020learning} for summarization tasks and the UltraChat dataset \cite{ding2023enhancing} for dialogue generation. Our aim is to demonstrate that high-quality demonstration data can be leveraged to construct synthetic preference datasets, which in turn can significantly improve both reward models and policy models without the sole reliance on human-annotated preferences. The experimental results show that our approach not only enhances model performance in terms of alignment with human judgments but also achieves reward learning without explicit human preferences.

\subsection{Experiments on the TL;DR Dataset}

In this experiment, we aim to train a language model for text summarization tasks using the TL;DR dataset~\cite{stiennon2020learning}, available on Hugging Face. We use all prompts from the TL;DR dataset as our prompt dataset. To create a demonstration dataset, we generate 10,000 high-quality summaries using a 6.9B parameter Pythia checkpoint~\cite{huang2024n+} that was trained via a RLHF pipeline with human-annotated preference data. This model is publicly available on Hugging Face\footnote{\url{https://huggingface.co/vwxyzjn}}.

In our IRL pipeline which iteratively updates the policy and the reward models through utilizing the demonstration data, 
we begin by performing supervised fine-tuning (SFT) on a pretrained 1B parameter Pythia model using the generated demonstration dataset, resulting in our initial SFT model. At each iteration, we construct a preference dataset by labeling the summaries generated from the 6.9B PPO-trained checkpoint as preferred and the outputs generated by our current 1B Pythia model as non-preferred. Using this preference dataset, we train a reward model initialized from our 1B SFT model. We then apply the Proximal Policy Optimization (PPO) algorithm, guided by the estimated reward model, to further fine-tune our 1B Pythia model, enhancing its performance beyond the initial SFT checkpoint. For the PPO algorithm setup, we follow the hyperparameters and experimental pipeline detailed in \cite{huang2024n+}. Our implementation is consistent with their codebase\footnote{\url{https://github.com/vwxyzjn/summarize_from_feedback_details}}.

We present our numerical results in Figure~\ref{figure:tl;dr}, showcasing the performance of the proposed iterative RLHF pipeline from three perspectives: (1) reward model accuracy, (2) reward scores measured by a 6.9B ground-truth reward model~\cite{huang2024n+} trained on the human-annotated TL;DR preference dataset, and (3) generating continuations from the prompt in the test dataset of the TL;DR dataset and then evaluate the win rates by GPT-4o to compare the text summarizations generated by IRL policy models and the 6.9B PPO-trained checkpoint.

Figure~\ref{figure:tl;dr} (a) presents the accuracy of our estimated 1B reward model on a human-annotated preference dataset of TL;DR\footnote{\url{https://huggingface.co/datasets/openai/summarize_from_feedback}}, which serves as a hold-out, out-of-distribution dataset since it is not used during our reward learning process. As shown in the figure, our IRL algorithm improves the reward model's accuracy over successive iterations, indicating better alignment with human preferences. Figure~\ref{figure:tl;dr} (b) illustrates the ground-truth reward scores assigned by the 6.9B Pythia reward model to summaries generated by our 1B policy model. The results indicate that our iterative RLHF method enhances the model's performance over iterations, as reflected by increasing reward scores. Additionally, Figure~\ref{figure:tl;dr} (c) presents the win rate of our 1B policy model compared to the 6.9B PPO-trained checkpoint, as evaluated by GPT-4. The iterative RLHF pipeline increases the SFT model's win rate from 24\% to 33\%, signifying that the proposed IRL method significantly outperforms the original SFT pipeline when learning from high-quality demonstrations. 

\begin{figure}[t]
    \centering
    \subcaptionbox{Reward Accuracy\label{fig:reward-accuracy}}{
        \includegraphics[width=0.31\textwidth]{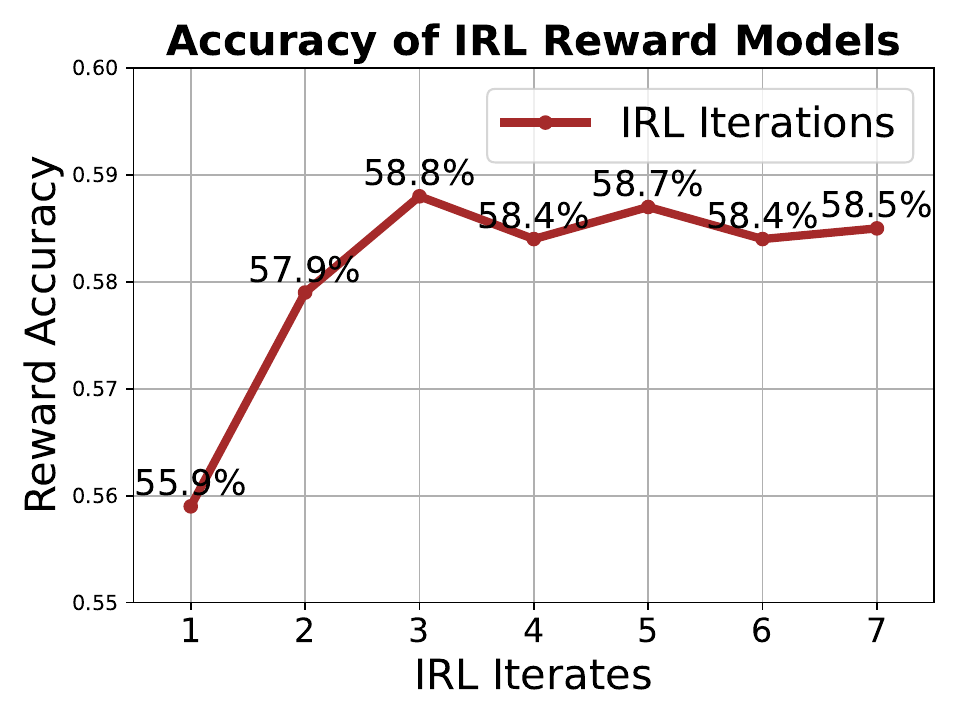}
    }
    \subcaptionbox{Ground-Truth Reward Score\label{fig:ground-truth}}{
        \includegraphics[width=0.31\textwidth]{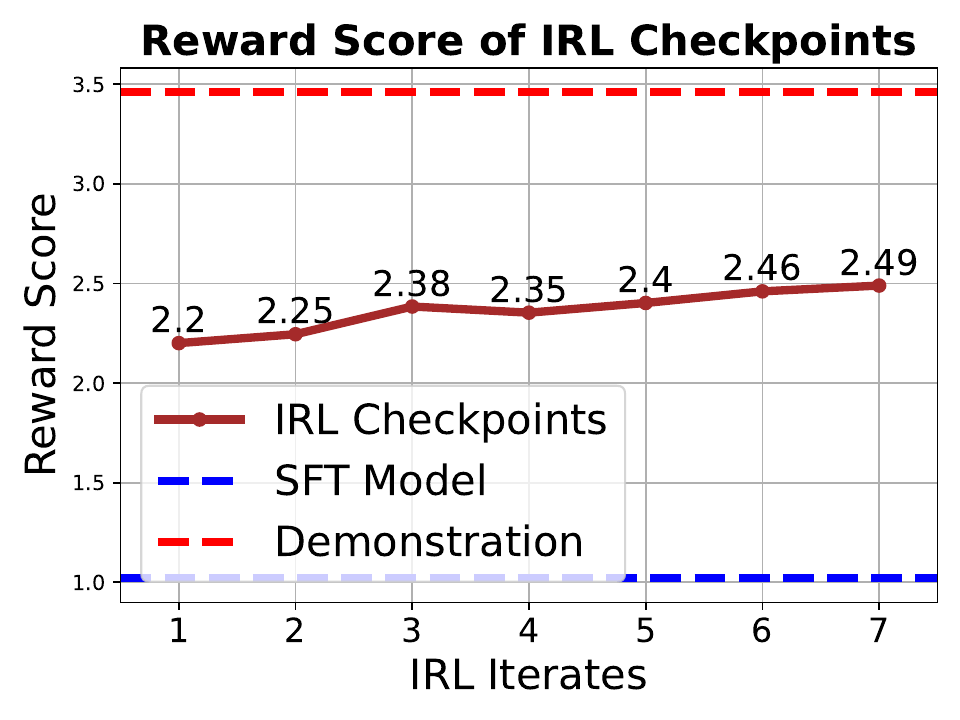}
    }
    \subcaptionbox{Win Rate by GPT-4o\label{fig:win-rate}}{
        \includegraphics[width=0.31\textwidth]{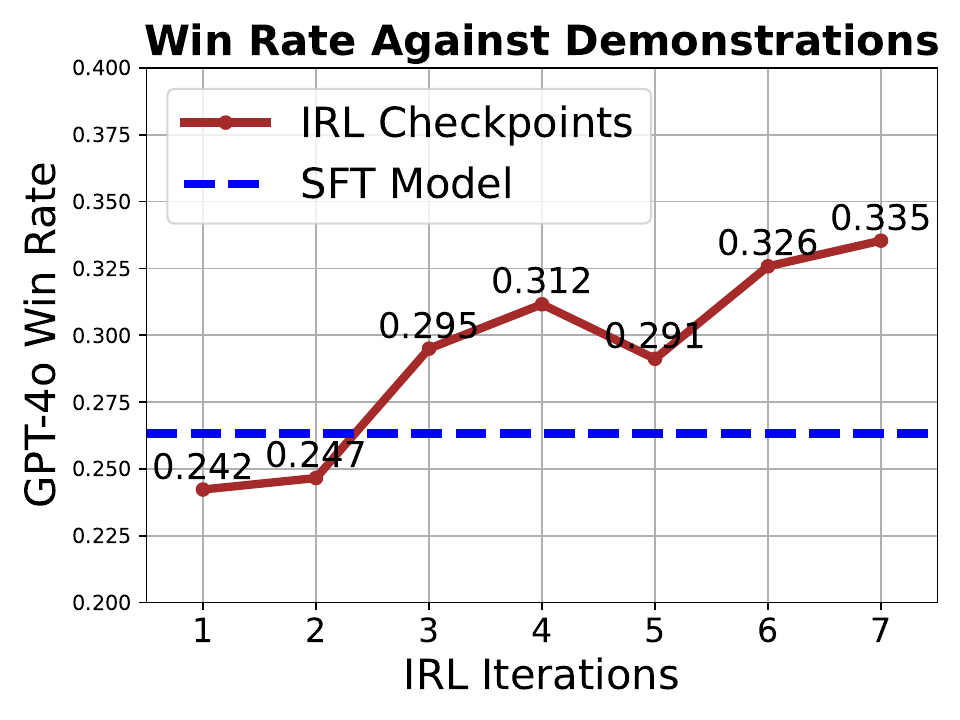}
    }
    \caption{Numerical Results of IRL Iterations with high-quality SFT data.}
    \label{figure:tl;dr}
\end{figure}

These findings collectively suggest that our approach effectively leverages high-quality demonstration data to construct meaningful preference datasets, leading to improvements in both the reward model and the policy model. The IRL pipeline not only enhances the alignment with human judgments but also achieves performance gains that are notable given the smaller size of the 1B parameter model compared to the 6.9B parameter baseline. By following this methodology, we demonstrate that even smaller models can achieve significant performance improvements through iterative RLHF processes towards imitating one larger language model when high-quality demonstrations are available. 

\subsection{Experiments on the UltraChat Dataset}

\begin{figure}[t]
    \centering
    \begin{subfigure}[b]{0.48\linewidth} 
        \includegraphics[width=\linewidth]{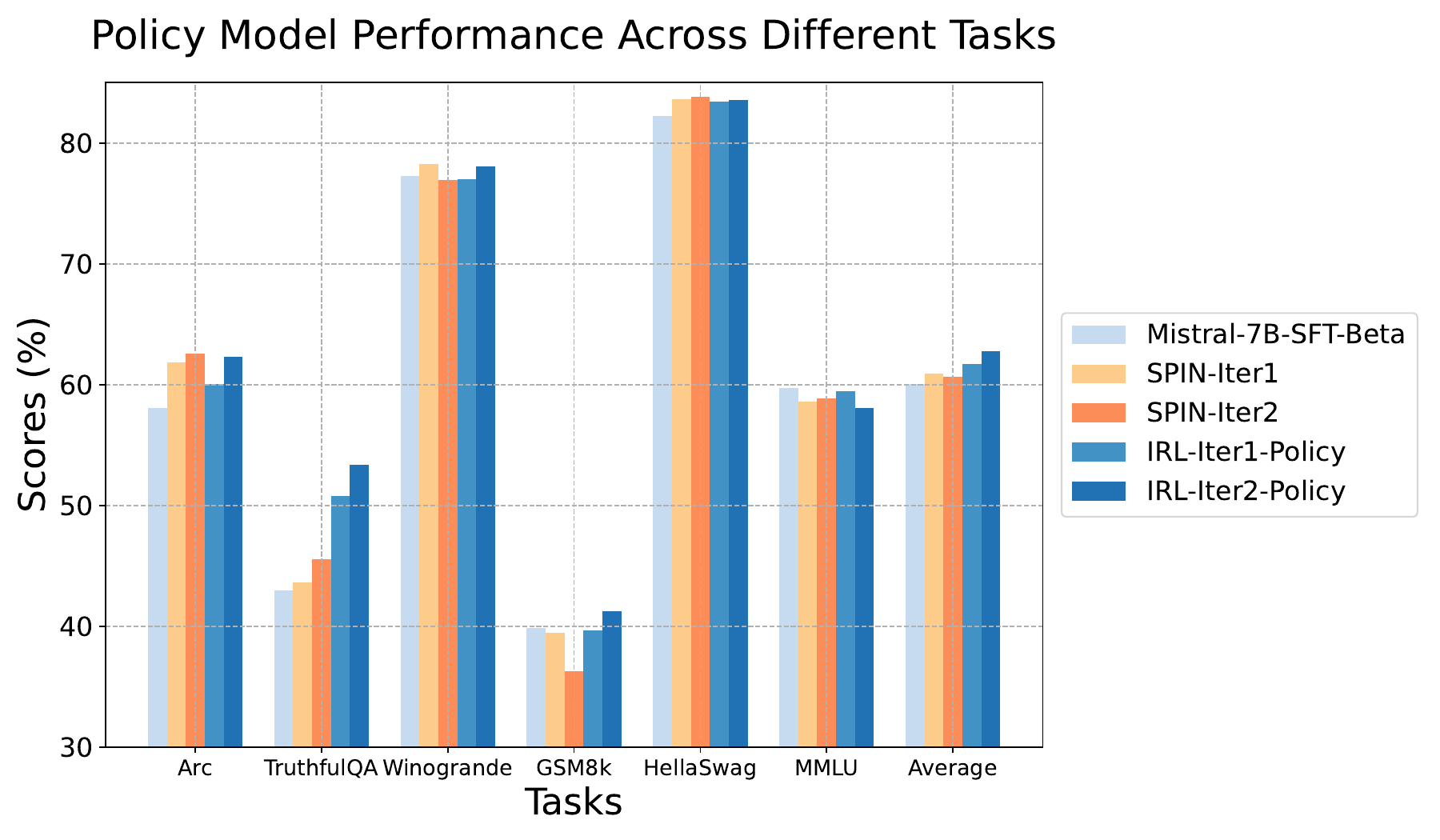}
        \vspace{-0.5cm} 
        \label{fig:llm_leaderboard}
    \end{subfigure}
    \hfill 
    \begin{subfigure}[b]{0.48\linewidth} 
        \includegraphics[width=\linewidth]{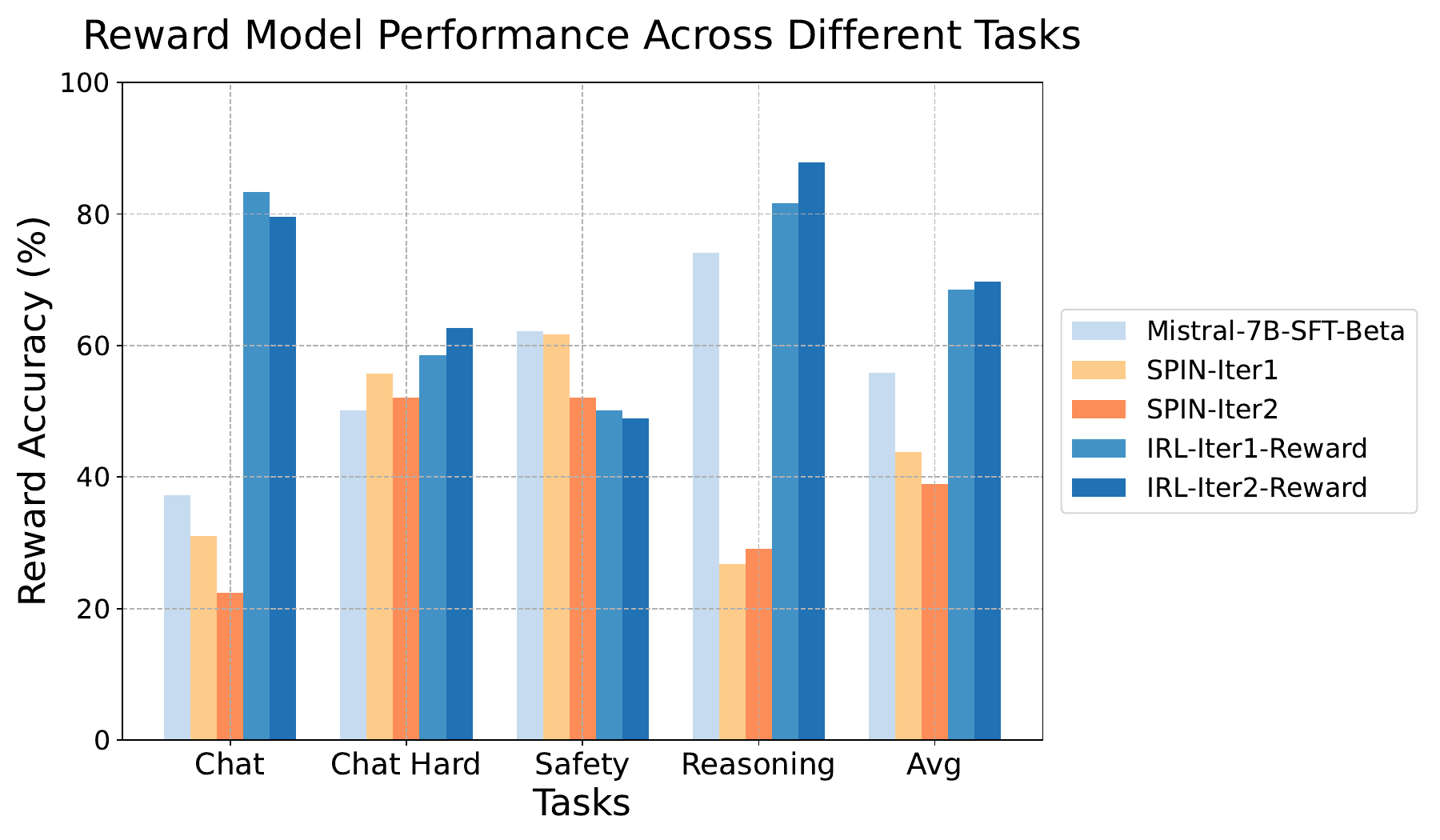}
        \vspace{-0.5cm}
        \label{fig:reward_bench}
    \end{subfigure}
    
    \caption{Overall Evaluation Results: (a) Policy Models on Open LLM Leaderboard; (b) Reward Models on Reward Bench.}
    \label{figure:merged_results}
\end{figure}

In this section, we present experiments demonstrating our proposed method applied to the UltraChat dataset~\cite{ding2023enhancing}, a high-quality dialogue dataset. For our experiments, we initialize both policy model and reward model from the checkpoint \texttt{HuggingFaceH4/mistral-7b-sft-beta}\footnote{\url{https://huggingface.co/HuggingFaceH4/mistral-7b-sft-beta}}.

\begin{table*}[ht]
    \centering
    \resizebox{0.99\textwidth}{!}{%
    \begin{tabular}{c | c c c c c c c}
    \toprule
        Tasks & Arc Challenge & TruthfulQA MC2 & Winogrande & GSM8k & HellaSwag & MMLU & Avg \\
        \# few shot & 25 & 0 & 5 & 5 & 10 & 5 & \\
        Metrics & acc\_norm & acc & acc & strict-match & acc\_norm & acc & \\
        \midrule
        \texttt{mistral-7b-sft-beta} & 58.10\% & 42.96\% & 77.26\% & 39.88\% & 82.23\% & 59.72\% & 60.03\% \\
        \hline
        \texttt{SPIN-Iter1} & 62.54\% & 48.71\% & 76.32\% & 34.65\% & 83.11\% & 56.49\% & 60.30\% \\
        \texttt{SPIN-Iter2} & {\bf 62.97\%} & {\bf 53.37\%} & 77.03\% & 24.79\% & {\bf 84.11\%} & 56.74\% & 59.84\% \\
    \hline
        \texttt{IRL-Iter1-Policy} & 60.07\% & 50.77\% & 77.03\% & 39.65\% & 83.41\% & {\bf 59.44\%} & 61.73\% \\
        \texttt{IRL-Iter2-Policy} & 62.29\% & {\bf 53.34\%} & {\bf 78.06\%} & {\bf 41.24\%} & 83.56\% & 58.10\% & {\bf 62.76\%} \\
    \bottomrule
    \end{tabular}%
    }
    \caption{Performance of Policy Models in Open LLm Leaderboard.}
    \label{Table:LLM_Leaderboard_Results}
\end{table*}

\begin{table*}[t]
    \centering
    \begin{tabular}{c | c c c c c}
    \toprule
        Tasks & Chat & Chat Hard & Safety & Reasoning & Avg \\
        \midrule
        \texttt{mistral-7b-sft-beta} & 37.29\% & {50.11\%} & {\bf 62.15\%} & 74.08\% & 55.91\% \\
        \midrule
        \texttt{SPIN-Iter1} & 31.01\% & {55.7\%} & {61.66\%} & {26.83\%} & 43.8\% \\
        \texttt{SPIN-Iter2} & 22.35\% & {52.08\%} & {52.04\%} & {29.09\%} & 38.89\% \\
        \midrule
        \texttt{IRL-Iter1-Reward} & {\bf 83.38\%} & {58.55\%} & {50.16\%} & {81.7\%} & 68.45\% \\
        \texttt{IRL-Iter2-Reward} & 79.61\% & {\bf 62.72\%} & {48.92\%} & {\bf 87.88\%} & {\bf 69.78\%} \\
    \bottomrule
    \end{tabular}%
    \caption{Performance of Reward Models in Reward-Bench.}
    \vspace{-0.3cm}
    \label{Table:Reward_Bench_Results}
\end{table*}

Since UltraChat is a supervised fine-tuning (SFT) dataset containing only demonstration data, we construct a synthetic preference dataset to train our reward model. Specifically, in the reward learning step for each IRL iteration, we treat the demonstration data from UltraChat as the preferred responses and the outputs generated by the IRL policy model as the rejected responses. This approach allows us to create preference pairs without requiring explicit human annotations. 

We evaluate our estimated reward models using the \texttt{allenai/reward-bench}\cite{lambert2024rewardbench}, assessing performance across various categories relevant to language understanding and generation. The results, illustrated in Figure \ref{figure:merged_results}, show that the reward model trained through the proposed IRL method achieves significant improvements compared to both the base model (initialized from the SFT model) and the implicit reward model extracted from the policy model trained using SPIN~\cite{chen2024self}. These findings indicate that high-quality demonstration datasets can effectively enhance reward models through leveraging IRL method which can construct synthetic preference pairs through pairing high-quality demonstrations and model generations.

\begin{table}[h]
    \centering
    \begin{tabular}{c | c c c}
    \toprule
        Tasks & First turn & Second turn & Average \\
        \midrule
        \texttt{mistral-7b-sft-beta} & 5.66 & 5.09 & 5.37 \\
        \midrule
        \texttt{SPIN-Iter1} & 6.75 & 5.56 & 6.16 \\
        \texttt{SPIN-Iter2} & 3.18 & 3.41 & 3.29 \\
        \midrule
        \texttt{IRL-Iter1-Policy} & 6.71 & 5.96 & 6.33 \\
        \texttt{IRL-Iter2-Policy} & \textbf{7.01} & \textbf{6.19} & \textbf{6.60} \\
    \bottomrule
    \end{tabular}%
    \caption{Performance of Policy Models in MT-Bench.}
    \label{Table:MT-Bench}
\end{table}

In each iteration of our IRL process, after updating the reward model, we fine-tune the previous IRL policy checkpoints using policy optimization methods guided by the estimated reward model. For the policy optimization subroutine, we follow the implementation details provided in the codebase of \cite{dong2024rlhf}\footnote{\url{https://github.com/RLHFlow/Online-RLHF}}. We employ the online DPO method as our policy trainer, which offers a memory-efficient approach for training large language models with limited computation resources.

To evaluate the effectiveness of our approach, we assessed our fine-tuned models on the Open LLM Leaderboard~\cite{eval-harness} and MT-Bench~\cite{zheng2023judging}. As shown in Fig~\ref{figure:merged_results}, Table~\ref{Table:LLM_Leaderboard_Results} and Table~\ref{Table:MT-Bench}, our method outperforms both the \texttt{HuggingFaceH4/mistral-7b-sft-beta} checkpoint and the SPIN method in Iterations 1 and 2, providing further evidence of the applicability and effectiveness of our approach. These results highlight the potential of leveraging high-quality demonstrations and synthetic preferences to enhance language model performance in dialogue generation tasks. 

By utilizing synthetic preference data derived from high-quality demonstrations, our approach effectively strengthens the reward model, which in turn enhances the policy model through iterative training. This strategy reduces the reliance on costly human-annotated preference data and demonstrates a scalable method for improving LLMs with high-quality demonstration dataset.

Through conducting extensive experiments comparing with the demonstration-only alignment baselines like SFT and SPIN, we demonstrate that our proposed IRL-based alignment method achieves the state-of-the-art performance when only demonstration data is available.


\section{Conclusion}
In this paper, we propose a new formulation for the alignment problem based on the IRL framework that utilizes only demonstration data. This approach enables us to simultaneously learn both a reward model and a policy model, resulting in a method that is more efficient than other demonstration-only methods, such as SFT. Our extensive experiments, on public reward benchmark and the Hugging Face Open LLM Leaderboard, demonstrate performance improvements over existing alignment baselines solely on demonstration data. These findings underscore that demonstration data offers greater utility than conventional wisdom suggests. As a future direction, we aim to integrate our proposed IRL-based methods with the current RLHF pipeline to achieve better flexibility and performance.

\bibliographystyle{IEEEtran} 
\bibliography{iclr2025_conference} 

\begin{thebibliography}{10}
\providecommand{\url}[1]{#1}
\csname url@samestyle\endcsname
\providecommand{\newblock}{\relax}
\providecommand{\bibinfo}[2]{#2}
\providecommand{\BIBentrySTDinterwordspacing}{\spaceskip=0pt\relax}
\providecommand{\BIBentryALTinterwordstretchfactor}{4}
\providecommand{\BIBentryALTinterwordspacing}{\spaceskip=\fontdimen2\font plus
\BIBentryALTinterwordstretchfactor\fontdimen3\font minus
  \fontdimen4\font\relax}
\providecommand{\BIBforeignlanguage}[2]{{%
\expandafter\ifx\csname l@#1\endcsname\relax
\typeout{** WARNING: IEEEtran.bst: No hyphenation pattern has been}%
\typeout{** loaded for the language `#1'. Using the pattern for}%
\typeout{** the default language instead.}%
\else
\language=\csname l@#1\endcsname
\fi
#2}}
\providecommand{\BIBdecl}{\relax}
\BIBdecl

\bibitem{achiam2023gpt}
OpenAI, ``Gpt-4 technical report,'' \emph{arXiv preprint arXiv:2303.08774},
  2023.

\bibitem{team2023gemini}
Gemini-Team, R.~Anil, S.~Borgeaud, Y.~Wu, J.-B. Alayrac, J.~Yu, R.~Soricut,
  J.~Schalkwyk, A.~M. Dai, A.~Hauth \emph{et~al.}, ``Gemini: a family of highly
  capable multimodal models,'' \emph{arXiv preprint arXiv:2312.11805}, 2023.

\bibitem{dubey2024llama}
A.~Dubey, A.~Jauhri, A.~Pandey, A.~Kadian, A.~Al-Dahle, A.~Letman, A.~Mathur,
  A.~Schelten, A.~Yang, A.~Fan \emph{et~al.}, ``The llama 3 herd of models,''
  \emph{arXiv preprint arXiv:2407.21783}, 2024.

\bibitem{geminiteam2024gemini15unlockingmultimodal}
Gemini-Team, ``Gemini 1.5: Unlocking multimodal understanding across millions
  of tokens of context,'' 2024.

\bibitem{ziegler2019fine}
D.~M. Ziegler, N.~Stiennon, J.~Wu, T.~B. Brown, A.~Radford, D.~Amodei,
  P.~Christiano, and G.~Irving, ``Fine-tuning language models from human
  preferences,'' \emph{arXiv preprint arXiv:1909.08593}, 2019.

\bibitem{ouyang2022training}
L.~Ouyang, J.~Wu, X.~Jiang, D.~Almeida, C.~Wainwright, P.~Mishkin, C.~Zhang,
  S.~Agarwal, K.~Slama, A.~Ray \emph{et~al.}, ``Training language models to
  follow instructions with human feedback,'' \emph{Advances in neural
  information processing systems}, vol.~35, pp. 27\,730--27\,744, 2022.

\bibitem{cen2025bridging}
Z.~Cen, Y.~Liu, S.~Zeng, P.~Chaudhari, H.~Rangwala, G.~Karypis, and R.~Fakoor,
  ``Bridging the training-inference gap in {LLM}s by leveraging self-generated
  tokens,'' \emph{Transactions on Machine Learning Research}, 2025.

\bibitem{zhao2023slichfsequencelikelihoodcalibration}
Y.~Zhao, R.~Joshi, T.~Liu, M.~Khalman, M.~Saleh, and P.~J. Liu, ``Slic-hf:
  Sequence likelihood calibration with human feedback,'' 2023.

\bibitem{an2023direct}
G.~An, J.~Lee, X.~Zuo, N.~Kosaka, K.-M. Kim, and H.~O. Song, ``Direct
  preference-based policy optimization without reward modeling,''
  \emph{Advances in Neural Information Processing Systems}, vol.~36, pp.
  70\,247--70\,266, 2023.

\bibitem{rafailov2024direct}
R.~Rafailov, A.~Sharma, E.~Mitchell, C.~D. Manning, S.~Ermon, and C.~Finn,
  ``Direct preference optimization: Your language model is secretly a reward
  model,'' \emph{Advances in Neural Information Processing Systems}, vol.~36,
  2024.

\bibitem{ethayarajh2024ktomodelalignmentprospect}
K.~Ethayarajh, W.~Xu, N.~Muennighoff, D.~Jurafsky, and D.~Kiela, ``Kto: Model
  alignment as prospect theoretic optimization,'' 2024.

\bibitem{ng2000algorithms}
A.~Y. Ng, S.~Russell \emph{et~al.}, ``Algorithms for inverse reinforcement
  learning.'' in \emph{Icml}, vol.~1, no.~2, 2000, p.~2.

\bibitem{ziebart2008maximum}
B.~D. Ziebart, A.~L. Maas, J.~A. Bagnell, A.~K. Dey \emph{et~al.}, ``Maximum
  entropy inverse reinforcement learning.'' in \emph{Aaai}, vol.~8.\hskip 1em
  plus 0.5em minus 0.4em\relax Chicago, IL, USA, 2008, pp. 1433--1438.

\bibitem{fu2017learning}
J.~Fu, K.~Luo, and S.~Levine, ``Learning robust rewards with adversarial
  inverse reinforcement learning,'' \emph{arXiv preprint arXiv:1710.11248},
  2017.

\bibitem{zeng2022maximum}
S.~Zeng, C.~Li, A.~Garcia, and M.~Hong, ``Maximum-likelihood inverse
  reinforcement learning with finite-time guarantees,'' \emph{Advances in
  Neural Information Processing Systems}, vol.~35, pp. 10\,122--10\,135, 2022.

\bibitem{chen2024self}
Z.~Chen, Y.~Deng, H.~Yuan, K.~Ji, and Q.~Gu, ``Self-play fine-tuning converts
  weak language models to strong language models,'' \emph{arXiv preprint
  arXiv:2401.01335}, 2024.

\bibitem{lambert2024rewardbench}
N.~Lambert, V.~Pyatkin, J.~Morrison, L.~Miranda, B.~Y. Lin, K.~Chandu,
  N.~Dziri, S.~Kumar, T.~Zick, Y.~Choi, N.~A. Smith, and H.~Hajishirzi,
  ``Rewardbench: Evaluating reward models for language modeling,'' 2024.

\bibitem{stiennon2020learning}
N.~Stiennon, L.~Ouyang, J.~Wu, D.~Ziegler, R.~Lowe, C.~Voss, A.~Radford,
  D.~Amodei, and P.~F. Christiano, ``Learning to summarize with human
  feedback,'' \emph{Advances in Neural Information Processing Systems},
  vol.~33, pp. 3008--3021, 2020.

\bibitem{huang2024n+}
S.~Huang, M.~Noukhovitch, A.~Hosseini, K.~Rasul, W.~Wang, and L.~Tunstall,
  ``The n+ implementation details of rlhf with ppo: A case study on tl; dr
  summarization,'' \emph{arXiv preprint arXiv:2403.17031}, 2024.

\bibitem{ding2023enhancing}
N.~Ding, Y.~Chen, B.~Xu, Y.~Qin, Z.~Zheng, S.~Hu, Z.~Liu, M.~Sun, and B.~Zhou,
  ``Enhancing chat language models by scaling high-quality instructional
  conversations,'' \emph{arXiv preprint arXiv:2305.14233}, 2023.

\bibitem{myrzakhan2024open}
A.~Myrzakhan, S.~M. Bsharat, and Z.~Shen, ``Open-llm-leaderboard: From
  multi-choice to open-style questions for llms evaluation, benchmark, and
  arena,'' \emph{arXiv preprint arXiv:2406.07545}, 2024.

\bibitem{zheng2023judging}
L.~Zheng, W.-L. Chiang, Y.~Sheng, S.~Zhuang, Z.~Wu, Y.~Zhuang, Z.~Lin, Z.~Li,
  D.~Li, E.~Xing \emph{et~al.}, ``Judging llm-as-a-judge with mt-bench and
  chatbot arena,'' \emph{Advances in Neural Information Processing Systems},
  vol.~36, pp. 46\,595--46\,623, 2023.

\bibitem{pomerleau1988alvinn}
D.~A. Pomerleau, ``Alvinn: An autonomous land vehicle in a neural network,''
  \emph{Advances in neural information processing systems}, vol.~1, 1988.

\bibitem{ross2010efficient}
S.~Ross and D.~Bagnell, ``Efficient reductions for imitation learning,'' in
  \emph{Proceedings of the thirteenth international conference on artificial
  intelligence and statistics}.\hskip 1em plus 0.5em minus 0.4em\relax JMLR
  Workshop and Conference Proceedings, 2010, pp. 661--668.

\bibitem{ross2011reduction}
S.~Ross, G.~Gordon, and D.~Bagnell, ``A reduction of imitation learning and
  structured prediction to no-regret online learning,'' in \emph{Proceedings of
  the fourteenth international conference on artificial intelligence and
  statistics}.\hskip 1em plus 0.5em minus 0.4em\relax JMLR Workshop and
  Conference Proceedings, 2011, pp. 627--635.

\bibitem{osa2018algorithmic}
T.~Osa, J.~Pajarinen, G.~Neumann, J.~A. Bagnell, P.~Abbeel, J.~Peters
  \emph{et~al.}, ``An algorithmic perspective on imitation learning,''
  \emph{Foundations and Trends{\textregistered} in Robotics}, vol.~7, no. 1-2,
  pp. 1--179, 2018.

\bibitem{ho2016generative}
J.~Ho and S.~Ermon, ``Generative adversarial imitation learning,''
  \emph{Advances in neural information processing systems}, vol.~29, 2016.

\bibitem{garg2021iq}
D.~Garg, S.~Chakraborty, C.~Cundy, J.~Song, and S.~Ermon, ``Iq-learn: Inverse
  soft-q learning for imitation,'' \emph{Advances in Neural Information
  Processing Systems}, vol.~34, pp. 4028--4039, 2021.

\bibitem{yu2017seqgan}
L.~Yu, W.~Zhang, J.~Wang, and Y.~Yu, ``Seqgan: Sequence generative adversarial
  nets with policy gradient,'' in \emph{Proceedings of the AAAI conference on
  artificial intelligence}, vol.~31, no.~1, 2017.

\bibitem{wu2021textgail}
Q.~Wu, L.~Li, and Z.~Yu, ``Textgail: Generative adversarial imitation learning
  for text generation,'' in \emph{Proceedings of the AAAI Conference on
  Artificial Intelligence}, vol.~35, no.~16, 2021, pp. 14\,067--14\,075.

\bibitem{ghosh2021helpful}
S.~Ghosh, Z.~Qi, S.~Chaturvedi, and S.~Srivastava, ``How helpful is inverse
  reinforcement learning for table-to-text generation?'' in \emph{Proceedings
  of the 59th Annual Meeting of the Association for Computational Linguistics
  and the 11th International Joint Conference on Natural Language Processing
  (Volume 2: Short Papers)}, 2021, pp. 71--79.

\bibitem{ghosh2021mapping}
S.~Ghosh and S.~Srivastava, ``Mapping language to programs using multiple
  reward components with inverse reinforcement learning,'' \emph{arXiv preprint
  arXiv:2110.00842}, 2021.

\bibitem{fu2022inverse}
Y.~Fu, D.~Xiong, and Y.~Dong, ``Inverse reinforcement learning for text
  summarization,'' \emph{arXiv preprint arXiv:2212.09917}, 2022.

\bibitem{li2024getting}
J.~Li, S.~Zeng, H.-T. Wai, C.~Li, A.~Garcia, and M.~Hong, ``Getting more juice
  out of the sft data: Reward learning from human demonstration improves sft
  for llm alignment,'' \emph{arXiv preprint arXiv:2405.17888}, 2024.

\bibitem{xu2024dpo}
S.~Xu, W.~Fu, J.~Gao, W.~Ye, W.~Liu, Z.~Mei, G.~Wang, C.~Yu, and Y.~Wu, ``Is
  dpo superior to ppo for llm alignment? a comprehensive study,'' \emph{arXiv
  preprint arXiv:2404.10719}, 2024.

\bibitem{ivison2024unpacking}
H.~Ivison, Y.~Wang, J.~Liu, Z.~Wu, V.~Pyatkin, N.~Lambert, N.~A. Smith,
  Y.~Choi, and H.~Hajishirzi, ``Unpacking dpo and ppo: Disentangling best
  practices for learning from preference feedback,'' \emph{arXiv preprint
  arXiv:2406.09279}, 2024.

\bibitem{cundysequencematch}
C.~Cundy and S.~Ermon, ``Sequencematch: Imitation learning for autoregressive
  sequence modelling with backtracking,'' in \emph{The Twelfth International
  Conference on Learning Representations}.

\bibitem{wulfmeier2024imitating}
M.~Wulfmeier, M.~Bloesch, N.~Vieillard, A.~Ahuja, J.~Bornschein, S.~Huang,
  A.~Sokolov, M.~Barnes, G.~Desjardins, A.~Bewley \emph{et~al.}, ``Imitating
  language via scalable inverse reinforcement learning,'' \emph{arXiv preprint
  arXiv:2409.01369}, 2024.

\bibitem{sun2024supervised}
H.~Sun, ``Supervised fine-tuning as inverse reinforcement learning,''
  \emph{arXiv preprint arXiv:2403.12017}, 2024.

\bibitem{christiano2017deep}
P.~F. Christiano, J.~Leike, T.~Brown, M.~Martic, S.~Legg, and D.~Amodei, ``Deep
  reinforcement learning from human preferences,'' \emph{Advances in neural
  information processing systems}, vol.~30, 2017.

\bibitem{bradley1952rank}
R.~A. Bradley and M.~E. Terry, ``Rank analysis of incomplete block designs: I.
  the method of paired comparisons,'' \emph{Biometrika}, vol.~39, no. 3/4, pp.
  324--345, 1952.

\bibitem{schulman2017proximal}
J.~Schulman, F.~Wolski, P.~Dhariwal, A.~Radford, and O.~Klimov, ``Proximal
  policy optimization algorithms,'' \emph{arXiv preprint arXiv:1707.06347},
  2017.

\bibitem{ahmadian2024back}
A.~Ahmadian, C.~Cremer, M.~Gall{\'e}, M.~Fadaee, J.~Kreutzer, A.~{\"U}st{\"u}n,
  and S.~Hooker, ``Back to basics: Revisiting reinforce style optimization for
  learning from human feedback in llms,'' \emph{arXiv preprint
  arXiv:2402.14740}, 2024.

\bibitem{li2023remax}
Z.~Li, T.~Xu, Y.~Zhang, Z.~Lin, Y.~Yu, R.~Sun, and Z.-Q. Luo, ``Remax: A
  simple, effective, and efficient reinforcement learning method for aligning
  large language models,'' in \emph{Forty-first International Conference on
  Machine Learning}, 2023.

\bibitem{dong2023raft}
H.~Dong, W.~Xiong, D.~Goyal, Y.~Zhang, W.~Chow, R.~Pan, S.~Diao, J.~Zhang,
  K.~Shum, and T.~Zhang, ``Raft: Reward ranked finetuning for generative
  foundation model alignment,'' \emph{arXiv preprint arXiv:2304.06767}, 2023.

\bibitem{ziebart2013principle}
B.~D. Ziebart, J.~A. Bagnell, and A.~K. Dey, ``The principle of maximum causal
  entropy for estimating interacting processes,'' \emph{IEEE Transactions on
  Information Theory}, vol.~59, no.~4, pp. 1966--1980, 2013.

\bibitem{zeng2022structural}
S.~Zeng, M.~Hong, and A.~Garcia, ``Structural estimation of markov decision
  processes in high-dimensional state space with finite-time guarantees,''
  \emph{arXiv preprint arXiv:2210.01282}, 2022.

\bibitem{cen2022fast}
S.~Cen, C.~Cheng, Y.~Chen, Y.~Wei, and Y.~Chi, ``Fast global convergence of
  natural policy gradient methods with entropy regularization,''
  \emph{Operations Research}, vol.~70, no.~4, pp. 2563--2578, 2022.

\bibitem{ji2024self}
X.~Ji, S.~Kulkarni, M.~Wang, and T.~Xie, ``Self-play with adversarial critic:
  Provable and scalable offline alignment for language models,'' \emph{arXiv
  preprint arXiv:2406.04274}, 2024.

\bibitem{dong2024rlhf}
H.~Dong, W.~Xiong, B.~Pang, H.~Wang, H.~Zhao, Y.~Zhou, N.~Jiang, D.~Sahoo,
  C.~Xiong, and T.~Zhang, ``Rlhf workflow: From reward modeling to online
  rlhf,'' \emph{arXiv preprint arXiv:2405.07863}, 2024.

\bibitem{eval-harness}
\BIBentryALTinterwordspacing
L.~Gao, J.~Tow, B.~Abbasi, S.~Biderman, S.~Black, A.~DiPofi, C.~Foster,
  L.~Golding, J.~Hsu, A.~Le~Noac'h, H.~Li, K.~McDonell, N.~Muennighoff,
  C.~Ociepa, J.~Phang, L.~Reynolds, H.~Schoelkopf, A.~Skowron, L.~Sutawika,
  E.~Tang, A.~Thite, B.~Wang, K.~Wang, and A.~Zou, ``A framework for few-shot
  language model evaluation,'' 12 2023. [Online]. Available:
  \url{https://zenodo.org/records/10256836}
\BIBentrySTDinterwordspacing

\end{thebibliography}

\clearpage

\section*{Appendix}

\section{Experiment Details}
In this section, we include the details in hyperparameters for our experiment in TL;DR and UltraChat. It is worth mentioning that we conducted a minimal random hyper-parameters search for the experiments in this paper and we mostly follow standard and readily available settings for the experiments whenever it is applicable.

\subsection{Experiment on TL;DR for training pythia-1b}
In the experiment of TL;DR, we use the TL;DR dataset to train the model \emph{pythia-1} 
For the RL trainer we used in our IRL pipeline, we utilize the PPO trainer. Here, we include the hyper-parameters for both reward modeling and PPO trainer as below.

\begin{table}[h!]
\centering
\caption{Reward modeling hyperparameters}
\begin{tabular}{ll}
\toprule
\textbf{Hyperparameter} & \textbf{Default Value} \\
\midrule
Number of Train Epochs  & 1 \\
Optimizer               & AdamW ($\epsilon = 1e^{-5}$, lr $= 3e^{-6}$) \\
Scheduler               & Cosine \\
Batch Size              & 64 \\
\bottomrule
\end{tabular}
\end{table}

\begin{table}[h!]
\centering
\caption{PPO hyperparameters}
\begin{tabular}{ll}
\toprule
\textbf{Hyperparameter} & \textbf{Default Value} \\
\midrule
Optimizer & AdamW ($\epsilon = 1e^{-5}$, lr $= 3e^{-6}$) \\
Scheduler & Linear \\
Batch Size & 512 \\
$\beta$ (KL Penalty Coefficient for RLHF) & 0.05 \\
$\gamma$ (Discount Factor) & 1.0 \\
$\lambda$ (for GAE) & 0.95 \\
$N_{\text{mb}}$ Number of Mini-batches & 1 \\
$K$ (Number of PPO Update Iterations Per Epoch) & 4 \\
$\epsilon$ (PPO's Policy Clipping Coefficient) & 0.2 \\
$\hat{\epsilon}$ (Value Clipping Coefficient) & 0.2 \\
$c_1$ (Value Function Coefficient) & 0.1 \\
Value Function Loss Clipping & True \\
Sampling Temperature & 0.7 \\
\bottomrule
\end{tabular}
\end{table}

\subsection{Experiment on UltraChat for training HuggingFaceH4/mistral-7b-sft-beta}
In the experiment of utilizing the UltraChat dataset to finetune \emph{HuggingFaceH4/mistral-7b-sft-beta}, 
we utilize the online DPO algorithm as the RL trainer in our IRL pipeline. Here, we include the hyper-parameter details as below:
\begin{table}[h!]
\centering
\caption{Reward modeling hyperparameters}
\begin{tabular}{ll}
\toprule
\textbf{Hyperparameter} & \textbf{Default Value} \\
\midrule
Number of Train Epochs  & 1 \\
Optimizer               & AdamW ($\epsilon = 1e^{-5}$, lr $= 5e^{-6}$) \\
Scheduler               & Cosine \\
Batch Size              & 64 \\
\bottomrule
\end{tabular}
\end{table}

\begin{table}[h!]
\centering
\caption{Online DPO hyperparameters}
\begin{tabular}{ll}
\toprule
\textbf{Hyperparameter} & \textbf{Default Value} \\
\midrule
Optimizer               & AdamW ($\epsilon = 1e^{-5}$, lr $= 5e^{-7}$) \\
Scheduler               & Cosine \\
Batch Size              & 64 \\
KL coefficient          & 0.1 \\
Best-of-N               & 32   \\
Pair Selection Strategy & Max-Min \\
\bottomrule
\end{tabular}
\end{table}

{
\section{Proof of Lemma \ref{lemma:objective_concentration}}
\label{proof:surrogate_objective_approximation}
\begin{proof}
For the policy optimization problem defined in \cref{def:inner_problem}, one can show that $\pi^*_{r_{\theta}}$ is the following closed-form expression:
\begin{equation}
        \pi^*_{r_{\theta}}(\bm{y}|\bm{x})=\frac{\pi_{\mathrm{ref}}(\bm{y}|\bm{x})\exp\left( r(\bm{x},\bm{y};\theta)\right)}{\sum_{\tilde{y}}\pi_{\mathrm{ref}}(\tilde{\bm{y}}|\bm{x})\exp\left( r(\bm{x},\tilde{\bm{y}};\theta)\right)}, \quad \forall \bm{x}, \bm{y}.
    \end{equation}
Then we can re-write the likelihood objective $L(\theta)$ defined in \cref{ML:formulation} as below:
\begin{align}
    L(\theta) &= \mathbb{E}_{\bm{x} \sim \mu( \cdot), \bm{y} \sim \pi^{\rm E}(\cdot | \bm{x} ) } \bigg[ \log \pi^*_{r_{\theta}}(\bm{y}|\bm{x}) \bigg] \nonumber \\
    &= \mathbb{E}_{\bm{x} \sim \mu( \cdot), \bm{y} \sim \pi^{\rm E}(\cdot | \bm{x} )} \bigg[ \log \bigg(  \frac{\pi_{\mathrm{ref}}(\bm{y}|\bm{x})\exp\left( r(\bm{x},\bm{y};\theta)\right)}{\sum_{\tilde{y}}\pi_{\mathrm{ref}}(\tilde{\bm{y}}|\bm{x})\exp\left( r(\bm{x},\tilde{\bm{y}};\theta)\right)}  \bigg) \bigg]  \nonumber \\
    &= \mathbb{E}_{\bm{x} \sim \mu( \cdot), \bm{y} \sim \pi^{\rm E}(\cdot | \bm{x} )} \bigg[   \log \Big( \pi_{\mathrm{ref}}(\bm{y}|\bm{x})\exp\big( r(\bm{x},\bm{y};\theta)\big) \Big)  - \log \Big( \sum_{\tilde{y}}\pi_{\mathrm{ref}}(\tilde{\bm{y}}|\bm{x})\exp\left( r(\bm{x},\tilde{\bm{y}};\theta)\right) \Big)  \bigg]  \nonumber \\
    &= 
    \mathbb{E}_{\bm{x} \sim \mu( \cdot), \bm{y} \sim \pi^{\rm E}(\cdot | \bm{x} ) } \bigg[ r(\bm{x}, \bm{y}; \theta) + \log \pi_{\rm ref}(\bm{y}|\bm{x}) \bigg] - \mathbb{E}_{\bm{x} \sim \mu( \cdot), \bm{y} \sim \pi^*_{r_{\theta}}(\cdot | \bm{x} )} \bigg[ r(\bm{x}, \bm{y}; \theta) - {D}_{\rm KL}\Big( \pi^*_{r_{\theta}}(\cdot | \bm{x}) \| \pi_{\text{ref}}(\cdot | \bm{x}) \Big) \bigg].  \nonumber
\end{align}
Moreover, given a dataset of collected expert trajectories, we have defined the estimation problem $\widehat{L}(\theta;\mathcal{D})$ as below:
{\small
\begin{align}
    \widehat{L}(\theta;\mathcal{D}) = \mathbb{E}_{(\bm{x}, \bm{y}) \sim \mathcal{D}} \big[ r(\bm{x}, \bm{y}; \theta) + \log \pi_{\rm ref}(\bm{y}|\bm{x}) \big] - \mathbb{E}_{\bm{x} \sim \mu( \cdot), \bm{y} \sim \pi^*_{r_{\theta}}(\cdot | \bm{x} )} \bigg[ r(\bm{x}, \bm{y}; \theta) - {D}_{\rm KL}\Big( \pi^*_{r_{\theta}}(\cdot | \bm{x}) \| \pi_{\text{ref}}(\cdot | \bm{x}) \Big) \bigg].   \nonumber
\end{align}
}
Then we have the following result:
\begin{align}
    |L(\theta) - \widehat{L}(\theta;\mathcal{D})| = \bigg| \mathbb{E}_{\bm{x} \sim \mu( \cdot), \bm{y} \sim \pi^{\rm E}(\cdot | \bm{x} ) } \bigg[ r(\bm{x}, \bm{y}; \theta) + \log \pi_{\rm ref}(\bm{y}|\bm{x}) \bigg] - \mathbb{E}_{(\bm{x}, \bm{y}) \sim \mathcal{D}} \big[ r(\bm{x}, \bm{y}; \theta) + \log \pi_{\rm ref}(\bm{y}|\bm{x}) \big] \bigg|. \nonumber
\end{align}
According to Assumption \ref{assumption:bound_reward}, we obtain that $0 \leq r(\bm{x}, \bm{y}; \theta) \leq C_r$ and $ C_p \leq \log \pi_{\rm ref}(\bm{y}|\bm{x}) < 0 $. Then by applying Hoeffding’s inequality, for any $\epsilon > 0$, we have the following result:
{\small
\begin{align}
    P\bigg(\bigg| \mathbb{E}_{\bm{x} \sim \mu( \cdot), \bm{y} \sim \pi^{\rm E}(\cdot | \bm{x} ) } \bigg[ r(\bm{x}, \bm{y}; \theta) + \log \pi_{\rm ref}(\bm{y}|\bm{x}) \bigg] - \mathbb{E}_{(\bm{x}, \bm{y}) \sim \mathcal{D}} \big[ r(\bm{x}, \bm{y}; \theta) + \log \pi_{\rm ref}(\bm{y}|\bm{x}) \big] \bigg| \geq \epsilon \bigg) \leq 2\exp \Big( -\frac{2|\mathcal{D}|\epsilon^2}{(C_r - C_p)^2} \Big). \nonumber
\end{align}
}
Then by setting $\delta = 2\exp \Big( -\frac{2|\mathcal{D}|\epsilon^2}{(C_r - C_p)^2} \Big) $, with probability greater than $1 - \delta$, we have
{\small
    \begin{align}
    &\bigg| \mathbb{E}_{\bm{x} \sim \mu( \cdot), \bm{y} \sim \pi^{\rm E}(\cdot | \bm{x} ) } \bigg[ r(\bm{x}, \bm{y}; \theta) + \log \pi_{\rm ref}(\bm{y}|\bm{x}) \bigg] - \mathbb{E}_{(\bm{x}, \bm{y}) \sim \mathcal{D}} \big[ r(\bm{x}, \bm{y}; \theta) + \log \pi_{\rm ref}(\bm{y}|\bm{x}) \big] \bigg| \leq ( C_r - C_p) \sqrt{\frac{\ln(2/\delta)}{2|\mathcal{D}|}},   \label{hoeffding:cumulative_reward}
\end{align}
}
where $C_r$ and $C_p$ is the constant defined in Assumption \ref{assumption:bound_reward}.
According to \cref{hoeffding:cumulative_reward}, we obtain the concentration bound to quantify the approximation between $L(\theta)$ and $\widehat{L}(\theta; \mathcal{D})$ as below:
\begin{align}
    |L(\theta) - \widehat{L}(\theta;\mathcal{D})| \leq (C_r - C_p) \sqrt{\frac{\ln(2/\delta)}{2|\mathcal{D}|}}, \quad \text{with probability greater than $1 - \delta$}.  \nonumber
\end{align}
This completes the proof of this lemma.
\end{proof}
}

\section{Proof of Lemma \ref{lemma:outer_gradient}}
\begin{proof}
In the surrogate estimation problem $\widehat{L}(\theta; \mathcal{D})$ defined in \cref{ML:estimation:surrogate}, the policy $\pi^*_{r_{\theta}}$ corresponds to the solution of the policy optimization problem \cref{def:inner_problem}. One can show that $\pi^*_{r_{\theta}}$ is the following closed-form expression:
\begin{equation}\label{eq:closed_form_lower}
        \pi^*_{r_{\theta}}(\bm{y}|\bm{x})=\frac{\pi_{\mathrm{ref}}(\bm{y}|\bm{x})\exp\left( r(\bm{x},\bm{y};\theta)\right)}{\sum_{\tilde{y}}\pi_{\mathrm{ref}}(\tilde{\bm{y}}|\bm{x})\exp\left( r(\bm{x},\tilde{\bm{y}};\theta)\right)}, \quad \forall \bm{x}, \bm{y}.
    \end{equation}
    Plugging \cref{eq:closed_form_lower} into \cref{ML:estimation:surrogate}, we obtain:
    \begin{equation}\label{eq:inverse_rl_single_level_appendix}
    \max _{\theta} \widehat{L}(\theta;\mathcal{D})=\mathbb{E}_{(\bm{x}, \bm{y})\sim\mathcal{D}}\left[ r(\bm{x},\bm{y};\theta) + \log \pi_{\rm ref}(\bm{y}|\bm{x}) \right] - \mathbb{E}_{\bm{x} \sim \mu} \big[
    \log\big(\sum_{\tilde{\bm{y}}}\pi_{\mathrm{ref}}(\tilde{\bm{y}}|\bm{x})\exp\left( r(\bm{x},\tilde{\bm{y}};\theta)\right)\big)\big]
    \end{equation}
    
    Calculating the derivative we get
    \begin{equation*}
    \begin{aligned}
    \max _{\theta} \widehat{L}(\theta;\mathcal{D}) =& \mathbb{E}_{(\bm{x}, \bm{y})\sim\mathcal{D}} [\nabla_{\theta}r(x,y;\theta)] - \mathbb{E}_{\bm{x} \sim \mu} \left[ \nabla_{\theta} \log\big(\sum_{\tilde{\bm{y}}}\pi_{\mathrm{ref}}(\tilde{\bm{y}}|\bm{x})\exp\left( r(\bm{x},\tilde{\bm{y}};\theta)\right)\big)\ \right]\\
    =& \mathbb{E}_{(\bm{x}, \bm{y})\sim\mathcal{D}} [\nabla_{\theta}r(x,y;\theta)] - \mathbb{E}_{\bm{x} \sim \mu} \left[\sum_{\bm{y}}\frac{\pi_{\mathrm{ref}}(\bm{y}|\bm{x})\exp\left(r(\bm{x},\bm{y};\theta)\right)}{\sum_{\tilde{\bm{y}}}\pi_{\mathrm{ref}}(\tilde{\bm{y}}|\bm{x})\exp\left( r(\bm{x},\tilde{\bm{y}};\theta)\right)} \nabla_{\theta} r(\bm{x},\bm{y};\theta) \right] \\
    =& \mathbb{E}_{(\bm{x}, \bm{y})\sim\mathcal{D}} [\nabla_{\bth}r(\bm{x},\bm{y};\theta)] - \mathbb{E}_{\bm{x} \sim \mu, \bm{y} \sim \pi^*_{\theta}(\cdot|\bm{x})} [\nabla_{\theta}r(\bm{x},\bm{y};\theta)].
    \end{aligned}
    \end{equation*}
    The proof is completed.
\end{proof}

\end{document}